\documentclass[letterpaper]{article} 
\usepackage[submission]{aaai25}  
\usepackage{times}  
\usepackage{helvet}  
\usepackage{courier}  
\usepackage[hyphens]{url}  
\usepackage{graphicx} 
\usepackage{natbib}  
\usepackage{caption} 
\usepackage{algorithm}
\usepackage{algorithmic}
\usepackage{booktabs} 
\usepackage{newfloat}
\usepackage{listings}
\usepackage{xcolor}
\usepackage{soul}
\usepackage{nameref}

\DeclareCaptionStyle{ruled}{labelfont=normalfont,labelsep=colon,strut=off} 
\floatstyle{ruled}
\newfloat{listing}{tb}{lst}{}
\floatname{listing}{Listing}
\title{Non-instructional Fine-tuning: Enabling Instruction-Following Capabilities in Pre-trained Language Models without Instruction-Following Data}
\author {
    Juncheng Xie, Shensian Syu, Hung-yi Lee
}
\affiliations{
    f07942150@ntu.edu.tw, d07921013@ntu.edu.tw, hungyilee@ntu.edu.tw
}

\usepackage{bibentry}

\begin{document}

\maketitle

\begin{abstract}
Instruction fine-tuning is crucial for today's large language models (LLMs) to learn to follow instructions and align with human preferences. Conventionally, supervised data, including the instruction and the correct response, is required for instruction fine-tuning. To obtain such data, some researchers prompted well-trained models like GPT-4 to generate instructions and correct responses. In this paper, we propose a novel approach that uses the first half of a random text from OpenWebText as the instruction and GPT-3.5-turbo or GPT-4-turbo to complete the text as the response. Despite the data being "non-instructional", we found that pre-trained LLMs fine-tuned on this data can gain instruction-following capabilities. This observation is verified by fine-tuning several well-known pre-trained LLMs (e.g., LLaMA-2-7B, LLaMA-3-8B, LLaMA-3-70B, Mistral-7B-v0.1). The "non-instructional data" also improved some models that underwent supervised fine-tuning and human preference alignment. Our LLaMA-3-70B-Instruct fine-tuned through "non-instructional data" is comparable with LLaMA-3.1-70B-Instruct on the Arena Hard leaderboard. We analyzed the "non-instructional data" and ensured it is devoid of content related to instruction fine-tuning. Our findings will inspire further investigation into how to develop instruction-following capabilities without explicit instruction-related data.
\end{abstract}

\section{Introduction} \label{intro}
In recent years, large language models (LLMs) like GPT-3 \citep{bai2022training} \citep{bai2022training} and LLAMA \citep{touvron2023llama} have showcased remarkable natural language processing capabilities across diverse domains \citep{zhao2021calibrate, wei2023zeroshot,wan2023gptre,sun2023text,li2023evaluating,gao2023exploring}. Previous studies have introduced instruction fine-tuning to align LLM training objectives with user goals. These methods involve either fine-tuning the model on various tasks using human-annotated prompts and feedback \citep{ouyang2022training}, or supervised fine-tuning utilizing public benchmarks and datasets augmented with manually or automatically generated instructions\citep{wang2022supernaturalinstructions}. Among these approaches, Self-Instruct tuning \citep{wang2023selfinstruct} stands out as a simple and effective method of aligning LLMs with human intent. This is achieved by learning from instruction-following data generated by state-of-the-art instruction-tuned teacher LLMs.

This paper finds that LLMs with instruction-following capabilities can be learned from "non-instructional data." In this context, "non-instructional data" refers to content that does not contain any explicit instructions. 
We employed publicly available datasets, such as OpenWebText \citep{radford_language_2019}, for ChatGPT\citep{openai2024gpt4} to continue writing. 
We demonstrate that data generated through distillation with continuous writing, even without explicit instructions, can enhance the capacity of LLMs to understand and execute tasks.
This paper investigates novel methodologies that empower LLMs to learn human instructions from a wider range of data, thus eliminating the need for manually annotated or explicitly generated instructional data.
Our contributions are summarized as follows:

\begin{enumerate}
  \item Introduce a simple framework for generating non-instructional datasets to finetune LLMs, enabling them to more effectively follow human instructions.
  
  \item Propose a methods for generating non-instructional data: conditional distillation and knowledge distillation with continuous writing.

  \item Propose a method of fine-tuning various LLMs using datasets generated by a novel approach. This method retains pre-fine-tuning scores on the Open LLM Leaderboard and significantly improves performance on the Arena Hard and MT Bench benchmark. Notably, our fine-tuned Meta-Llama-3-8b model demonstrated substantial gains on Arena Hard, compared to other strong SFT dataset, and the fine-tuned Meta-Llama-3-70b-Instruct model achieved the highest recorded score of 57.0, surpassing even the more advanced Meta-Llama-3.1-70b-Instruct. These results underscore the effectiveness of our fine-tuning approach in enhancing the instruction-following capabilities of large language models.

  \item Introduce the use of lora-base for model enhancement, demonstrating its effectiveness in improving performance. This technique involves merging the LoRA module fine-tuned on the foundation (base) model with the Instruct model, showcasing improvements across various benchmarks without additional training overhead.

  
\end{enumerate}
 
\section{Related Work} \label{related_work}

In the landscape of LLM distillation and fine-tuning, Stanford Alpaca \citep{alpaca} and Code Alpaca \citep{codealpaca} underscore the utility of leveraging the LLaMA model and GPT variants to generate instructional content, demonstrating the power of distillation for tailored instruction following. Extending this concept, Alpaca-CoT \citep{si2023empirical}, with its focus on instruction tuning, and Baize \citep{xu2023baize}, showcasing a novel self-chat method with ChatGPT for dialogue improvement, represent innovative strides in model enhancement. Vicuna \citep{vicuna2023} and Koala \citep{koala_blogpost_2023}, through user-shared conversations and web-collected dialogues, respectively, imply the indirect but significant influence from the distillation of high-caliber models like ChatGPT. \cite{peng2023instruction} elucidates the direct application of GPT-4 for knowledge distillation, enriching the field with nuanced methodologies. UltraChat \citep{ding2023enhancing} further contributes to this narrative by presenting a large-scale, diverse dataset of instructional conversations, aiming to elevate the performance of open-source models like UltraLLaMA beyond existing benchmarks such as Vicuna.

MAGPIE \citep{xu2024magpiealignmentdatasynthesis} introduces a scalable method to synthesize high-quality instruction data by leveraging the auto-regressive nature of aligned LLMs, such as LLaMA-3-Instruct, to self-generate instructions and responses without the need for human seed data or prompt engineering. By using this method, MAGPIE efficiently generates large-scale datasets that significantly enhance the performance of models fine-tuned on them, surpassing even officially aligned models like LLaMA-3-8B-Instruct on various alignment benchmarks. This positions MAGPIE as a critical advancement in the creation of alignment datasets, offering a robust alternative to traditional human-involved methods or synthetic data generation approaches that rely on extensive prompt engineering.

The methods mentioned above rely on instructional data to train LLMs. This paper represents the first attempt to explore the potential of acquiring instruction-following capabilities from non-instructional data.
 
\section{Non-instructional Dataset Generation}\label{dataset_generation}

\begin{figure*}[h]
\begin{center}
\includegraphics[width=1\textwidth]{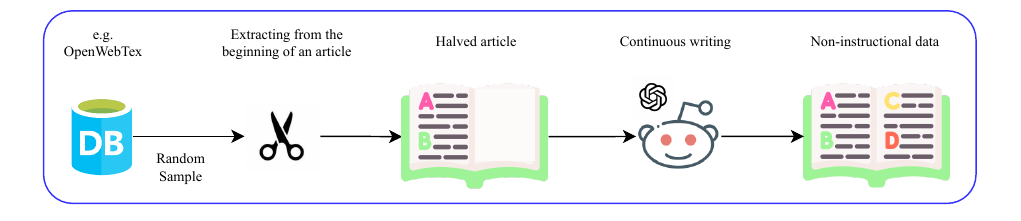}
\end{center}
\caption{ Our framework for distillation involves using a specific dataset to prompt ChatGPT for continued writing, simulating a targeted context. }
\label{framework}
\end{figure*}

In contrast to the intricate process traditionally involved in creating instruction-following datasets, which includes amassing extensive texts, categorizing them, and further formatting these texts into instruction-based dialogues, our methodology offers a streamlined alternative. We bypass the multifaceted stages of traditional data preparation, including the generation of instruction-formatted content, by directly distilling knowledge from substantial models like ChatGPT. Our framework is shown in Figure \ref{framework}. In the following sections, we refer to our datasets as \textbf{non-instructional datasets}.

\subsection{Distillation from OpenAI LLMs} \label{conditional_distillation}

Our approach employs 'halving and completion' on 80,000 pieces of data uniformly sampled from the OpenWebText\citep{radford_language_2019} corpus sourced from Reddit. 
This method involves selecting a midpoint for halving that is uniformly sampled from between the first and the last quarter of the article (measured by word count), and then using an instruction-following LLM as the teacher model to (almost) deterministically generate the concluding half with a temperature of 0.0. 
Such a process leverages the inherent capabilities of LLMs to produce diverse and contextually rich content. 
Employing this kind of fine-tuning data mirrors continued unsupervised pretraining. 
We selected the latest and most cost-effective versions from OpenAI: gpt-3.5-turbo-0125 and gpt-4-0125-preview, as our teacher models for continuous writing. 
The outcomes of training on these distilled datasets across various models are elaborated in Section~\ref{conditional_eval}.

We will release the data under the terms and conditions of OpenAI's Terms of Use \footnote{https://openai.com/policies/terms-of-use/} and Usage Policies \footnote{https://openai.com/policies/usage-policies/}.

\subsection{Distillation from Anthropic AI LLMs} \label{other_conditional_distillation}

As discussed in Section \ref{benchmarks}, some of the benchmarks we used employ GPT-4 as the judge. However, based on validation and discussions in \citep{zheng2023judging,panickssery2024llm,alpaca_eval}, it has been noted that GPT-4 not only tends to favor its own generations but may also favor the generations of models fine-tuned on data distilled from itself. To demonstrate the robustness of our method under different strong LLMs as teacher models, we also performed distillation using Anthropic AI's Claude-3-Haiku, Claude-3-Sonnet, and Claude-3-Opus.

Due to time and budget constraints, we limited our distillation to 10,000 pieces of OpenWebText data. Apart from adding a “system prompt”: “Please continue directly from the end of the given sentence without repeating it,” the remaining settings were identical to those described in Section \ref{conditional_distillation}. This adjustment ensures that the process maintains consistency and leverages the same systematic approach used with OpenAI models, allowing for a fair comparison of the efficacy of distillation across different LLMs.

\section{Experiment Setup}
\label{exp_setup}

\subsection{LLMs} 
\label{LLms}

In our experiments, we utilize a diverse set of Large Language Models (LLMs) for fine-tuning with our dataset, alongside others for comparative analysis. The fine-tuning models include LLaMA-2-7B\citep{touvron2023llama}, Meta-Llama-3-8B and Meta-Llama-3-70B \citep{llama3modelcard}, which come in both foundation and Instruct (chat) variants. We also employ the Mistral-7B-v0.1 series\citep{jiang2023mistral}, which includes Mistral-7B-v0.1, Mistral-7B-Instruct-v0.1, and Mistral-7B-Instruct-v0.2.

\subsection{Finetuning Details} 
\label{ft_details}


The finetuning procedure adopted in our study is relatively straightforward, utilizing the codebase provided by LLaMA-Factory\footnote{https://github.com/hiyouga/LLaMA-Factory}\citep{llama-factory}, which is tailored for the efficient finetuning of large language models. We opt for a supervised finetuning (sft) mode, though the data format would traditionally align with a pretraining paradigm. The finetuning module chosen is LoRA, with the process spanning 3 epochs. Detailed information regarding the specific versions used, the computational hardware, and the finetuning commands are discussed further in appendix \ref{sec:finetuning_details}.
\section{Evaluation}
\label{conditional_eval}


\subsection{Benchmarks} \label{benchmarks}
To ensure a fair comparison of model capabilities, we select three benchmarks: MT-Bench \citep{zheng2023judging}, Open LLM Leaderboard \citep{open-llm-leaderboard}, and Arena Hard \citep{arenahard2024}.

\paragraph{MT-Bench}
MT-Bench evaluates LLMs' dialogue and instruction-following capabilities using 80 multi-turn questions. Each response is scored by GPT-4 (gpt-4-0613) on a scale from 1 to 10, with an average score calculated over three rounds. It uses FastChat\footnote{https://github.com/lm-sys/FastChat} for model inference, ensuring efficient analysis. The scores for subsequent experiments are averaged over three inference runs.



\paragraph{Open LLM Leaderboard}
The Open LLM Leaderboard\footnote{https://huggingface.co/spaces/HuggingFaceH4/open\_llm\_leaderboard} uses the Eleuther AI Language Model Evaluation Harness\footnote{https://github.com/EleutherAI/lm-evaluation-harness} to evaluate models across six benchmarks. This comprehensive assessment covers reasoning, knowledge, and truthfulness. Scores are benchmarked against those on the leaderboard website for consistency.

\paragraph{Arena Hard}
Arena Hard\footnote{https://github.com/lm-sys/arena-hard-auto} evaluates LLMs with 500 complex, real-world questions from Chatbot Arena\footnote{https://huggingface.co/spaces/lmsys/chatbot-arena-leaderboard}. Pairwise comparisons against a strong baseline (GPT-4-0314) using GPT-4-Turbo ensure robust assessment. This benchmark emphasizes real-world application, with high separability (87.4\%) and agreement with human preference rankings (89.1\%).

\subsection{IFEval Benchmark}

To further evaluate the instruction-following capabilities of our models, we include the IFEval dataset \citep{zhou2023instructionfollowingevaluationlargelanguage}, specifically tests models on their ability to follow pattern-level or syntax-level instructions. This benchmark focuses on \textit{verifiable instructions}—clear, objective directives such as "write in more than 400 words" or "mention the keyword of AI at least 3 times." Comprising 25 types of instructions and around 500 prompts, IFEval is designed to verify whether a model can adhere to precise instruction formats, including specific structural requirements. We include this benchmark in our evaluation to highlight the models' ability to comply with detailed instructional patterns.

\subsection{Results}



\begin{table*}[ht]
\centering
\footnotesize 
\begin{tabular}{lcccccc}
\toprule
ID & Backbone Model & Template & Fine-tuned Modules & Fine-tuning Data & MT Bench & OLL Score \\
\midrule
1 & llama2-7b & one\_shot & - & - & $3.88$ &  50.97 \\
2 & llama2-7b-chat & llama-2 & - & - & 6.27 &  50.74 \\
3 & llama2-7b & one\_shot & lora & OpenWebText 80k & $3.82$ & 50.65\\
4 & llama2-7b & one\_shot & lora & llama2-7b-chat 80k & $4.59$ & 49.49 \\
5 & llama2-7b & one\_shot & lora & gpt3.5-turbo 80k & $4.93$ & 50.64 \\
6 & llama2-7b-chat & llama-2 & lora-base & gpt3.5-turbo 80k & $6.60$ & 50.27 \\
7 & llama2-7b-chat & llama-2 & lora & gpt3.5-turbo 80k & $6.33$ & 51.75\\
8 & llama2-7b & one\_shot & lora & gpt4-turbo 80k & $5.12$ & 52.71\\
9 & llama2-7b-chat & llama-2 & lora-base & gpt4-turbo 80k & $ \textbf{6.71}$ & 49.86\\
10 & llama2-7b-chat & llama-2 & lora & gpt4-turbo 80k & $6.57$ & 52.98 \\
\cline{1-7}
11 & Mistral-7B-v0.1 & zephyr & - & - & $3.73$ & 60.97\\
12 & Mistral-7B-v0.1 & zephyr & lora & undistilled 80k & $3.57$ & 61.01\\
13 & Mistral-7B-v0.1 & zephyr & lora & gpt4-turbo 80k & $7.29$ & 62.00\\
14 & Mistral-7B-Instruct-v0.1 & mistral & - & - & 6.84 & 54.96\\
15 & Mistral-7B-Instruct-v0.1 & mistral & lora-base & gpt4-turbo 80k & $6.54$ & 54.78\\
16 & Mistral-7B-Instruct-v0.1 & mistral & lora & gpt4-turbo 80k & $7.02$ & 57.27 \\
17 & Mistral-7B-Instruct-v0.2 & mistral & - & - & 7.6 & 65.71\\
18 & Mistral-7B-Instruct-v0.2 & mistral & lora-base & gpt4-turbo 80k & $7.46$ & 60.82 \\
19 & Mistral-7B-Instruct-v0.2 & mistral & lora & gpt4-turbo 80k & $\textbf{7.74}$ & 62.51\\
\cline{1-7}
20 & Meta-Llama-3-8b & llama-3 & - & - & 5.5 & 62.62 \\
21 & Meta-Llama-3-8b-Instruct  & llama-3 & - & - & 7.86 & 66.87 \\
22 & Meta-Llama-3-8b  & llama-3 & lora & gpt4-turbo 80k & 7.03 & 63.67 \\
23 & Meta-Llama-3-8b-Instruct & llama-3 & lora & gpt4-turbo 80k & 7.97 & 64.88 \\
24 & Meta-Llama-3-8b-Instruct & llama-3 & lora-base & gpt4-turbo 80k & \textbf{8.21} & 60.10 \\
\cline{1-7}
25 & Meta-Llama-3-70b & llama-3 & - & - & 2.71 & 73.96 \\
26 & Meta-Llama-3-70b-Instruct  & llama-3 & - & - & 8.63 & 77.88 \\
27 & Meta-Llama-3-70b  & llama-3 & lora & gpt4-turbo 80k & 8.18 & -\\
28 & Meta-Llama-3-70b-Instruct & llama-3 & lora & gpt4-turbo 80k & \textbf{9.03} & -\\
29 & Meta-Llama-3-70b-Instruct & llama-3 & lora-base & gpt4-turbo 80k & 8.71 & -\\
\bottomrule
\end{tabular}
\caption{Performance of LLMs on MT-Bench. OLL Score refers to the Average Score on the Open LLM Leaderboard.}
\label{main-table-mt-bench}
\end{table*}

In Table~\ref{main-table-mt-bench}, the performance of LLMs on MT-Bench, aiming to align more closely with human preferences on open-ended questions, is evaluated through a series of experiments using various fine-tuning datasets. 
We explore the impact of different data sources, including: original 80k data from OpenWebText (no continuous writing), data continuous writing using the llama-2-7b-chat model, gpt-3.5-turbo-0125 and gpt-4-0125-preview. 
The 'Template' refers to the template used during MT-Bench inference, as detailed in the documentation found here\footnote{https://github.com/lm-sys/FastChat/blob/main/fastchat/conversation.py}. 
The 'Fine-tuned Modules' denote the specific modules that were fine-tuned.
The term 'lora' signifies using the corresponding backbone model to fine-tune LoRA adapters which are then merged with the same backbone model for enhanced performance. 
'lora-base' is unique to Instruct/chat models, indicating their merging with adapters fine-tuned on their corresponding foundation model.

The impact of different data sources on LLaMA-2-7B can be observed by comparing ID 3, 4, 5 and 8. 
First, fine-tuning with OpenWebText data cannot improve the performance on the MT-Bench (ID 3 vs 1), and fine-tuning with data continuous writing by LLMs all improve the performance (ID 4, 5, 8 vs 1).
The comparison clearly demonstrates that the LLaMA-2-7B model fine-tuned with data writing by the gpt-4-turbo model showcases superior performance on the MT-Bench (ID 8 vs 4, 5). 
This highlights the significance of selecting high-quality data writing by advanced models for fine-tuning purposes, leading us to primarily utilize the dataset from gpt-4-0125-preview as the cornerstone of this research. 

The effectiveness of non-instructional data not just for enhancing the foundation models but also for Instruct/chat variants.
Additionally, LLaMA-2-7B-chat, Meta-Llama-3-8b-Instruct and Meta-Llama-3-70b-Instruct show improved performance upon fine-tuning with the non-instructional datasets  (ID 9, 10 vs 2 and ID 23, 24 vs 21 and ID 28, 29 vs 26). 
Notably, even when chat/Instruct models are merged with LoRA modules fine-tuned on corresponding foundation models, an enhancement in performance is observed. 
This suggests that our LoRA modules possess a function akin to chat vectors\citep{huang2024chat}.

This indicates a significant improvement in dialogue capabilities across both model types.


Reflecting on the nuances of our findings, it becomes apparent that the three foundation models, which have only undergone pretraining, and Instruct/chat models exhibit significant performance improvements on the MT-Bench when fine-tuned with our non-instructional data. This observation underscores the efficacy of our fine-tuning approach, not just for the foundation models but also for those specifically designed for Instruct/chat interactions.

Moreover, an intriguing aspect of our analysis highlights that Instruct/chat models, even when merged with LoRA modules fine-tuned on corresponding foundation models, demonstrate enhanced performance. This outcome suggests that the LoRA modules can carry functionalities and add to other models, contributing to improved dialogue capabilities. Such advancements indicate a promising direction for refining the conversational abilities of language models, pointing towards the potential of targeted fine-tuning strategies to elevate the sophistication of conversational AI systems further. 

We will release the data under the terms and conditions of OpenAI's Terms of Use and Usage Policies.

\subsection{Claude distillation data}
Our experiments involving Claude-3 distillation data reveal significant improvements across various models, supporting our hypothesis that non-instructional data from powerful LLMs can be highly beneficial. The results in Table \ref{tab:claude_results} indicate that even smaller datasets distilled from Claude-3-Haiku, Claude-3-Sonnet, and Claude-3-Opus can effectively enhance the performance of models such as Mistral-7B-v0.1, Meta-Llama-3-8b, and Meta-Llama-3-70b.

When comparing these results to our previous evaluations using gpt-4-0125-preview. data, it is evident that the Claude-3 distillation data is equally impactful. For instance, the Meta-Llama-3-70b-Instruct model fine-tuned on Claude-3 datasets consistently achieved high MT Bench scores, with a peak score of 9.00 using Claude-3-Opus. This underscores the robustness and utility of leveraging high-quality non-instructional data from strong LLMs for fine-tuning, aligning with our findings that such data can significantly boost model performance without relying on traditional instruction-following datasets.

\begin{table}[ht]
\centering
\footnotesize
\begin{tabular}{lccc}
\toprule
Backbone  & Haiku & Sonnet & Opus \\
\midrule
Mistral-7B-v0.1 & 6.87 & 6.45 & 6.63 \\
Meta-Llama-3-8b & 5.51 & 6.16 & 5.56 \\
Meta-Llama-3-8b-Instruct & 7.62 & 7.73 & 7.91 \\
Meta-Llama-3-8b lora-base & 8.06 & 8.15 & 8.03 \\
Meta-Llama-3-70b & 7.38 & 7.50 & 7.65 \\
Meta-Llama-3-70b-Instruct & 8.99 & 8.91 & 8.80 \\
Meta-Llama-3-70b lora-base & 8.79 & 8.90 & \textbf{9.00} \\
\bottomrule
\end{tabular}
\caption{Results of Fine-Tuning on Claude Distillation Data}
\label{tab:claude_results}
\end{table}

\subsection{Comparison with alpaca data}

Alpaca \citep{alpaca}, developed by Stanford, is a well-known dataset for instruction fine-tuning based on the LLaMA model. It comprises 52,000 instruction-following demonstrations generated using OpenAI's text-davinci-003 model. 
In table \ref{table:comparison_alpaca}, we compare the performance of Mistral-7B-v0.1 fine-tuned on the original Alpaca data, the GPT-4 distilled Alpaca data \citep{peng2023instruction} \footnote{https://github.com/Instruction-Tuning-with-GPT-4/GPT-4-LLM} , and a smaller, non-instructional dataset generated from GPT-4-Turbo and Claude-3-Haiku. 
Our results show that models fine-tuned on just 10,000 examples from GPT-4-Turbo and Claude-3-Haiku outperform those trained on the larger Alpaca dataset, including the GPT-4 distilled version. Specifically, the MT Bench scores for the GPT-4-Turbo and Claude-3-Haiku datasets are 6.75 and 6.87, respectively, compared to 5.67 for the original Alpaca and 6.56 for the GPT-4 distilled Alpaca. This demonstrates that a smallernon-instructional dataset can achieve superior performance, highlighting the efficiency and potential of our approach in instruction-following tasks.

\begin{table}[ht]
\centering
\begin{tabular}{|c|c|}
\hline
\textbf{Dataset} & \textbf{MT Bench Score} \\ \hline
Alpaca (Original) & 5.67 \\ \hline
Alpaca (GPT-4 Distilled) & 6.56 \\ \hline
GPT-4-Turbo & 6.75 \\ \hline
Claude-3-Haiku & 6.87 \\ \hline
\end{tabular}
\caption{Comparison of Mistral-7B-v0.1 finetuned on different datasets and their MT Bench scores. The Alpaca datasets contain 52K instruction-following examples, while the GPT-4-Turbo and Claude-3-Haiku distillation non-instructional datasets each contain only 10K examples.}
\label{table:comparison_alpaca}
\end{table}

\subsection{Open LLM Leaderboard Results}

\begin{table*}[ht]
\centering
\begin{tabular}{cccccccc}
\toprule
ID & AVG & ARC & HellaSwag & MMLU & TruthfulQA & Winogrande & GSM-8K \\
\midrule
1 &  50.97 & 53.07 &     78.59 & 46.87 &      38.76 &      74.03 &  14.48 \\
 2 &  50.74 &  52.9 &     78.55 & 48.32 &      45.57 &      71.74 &   7.35 \\
 3 &              50.65 & 53.41 &     78.62 & 46.26 &      38.82 &      74.66 &  12.13 \\
 4 &  49.49 & 53.84 &     74.65 & 46.36 &      39.06 &      71.03 &  11.98 \\
 5 &  50.64 &     53.84 &    75.77 &   46.13 & 41.42 &       72.06 &   14.63 \\
 6 &              50.27 & 51.45 &     69.38 &  48.20 &      46.62 &       67.40 &  18.57 \\
 7 &  51.75 & 52.05 &     73.89 & 48.19 &      44.35 &      71.98 &  20.02 \\
 8 &              52.71 & 55.55 &     77.27 & 46.75 &      48.63 &      74.03 &  14.03 \\
 9 &  49.86 & 52.56 &     71.37 & 48.34 &      48.22 &      66.61 &  12.05 \\
10 &             52.98 & 54.78 &     74.63 & 48.79 &      48.45 &      72.85 &  18.35 \\
\bottomrule
\end{tabular}
\caption{
Performance on Open LLM Leaderboard Tasks. The IDs refer to the models in Table~\ref{main-table-mt-bench}. The full table is available in the Appendix (Table~\ref{full-table-oll}).
}
\label{main-table-oll}
\end{table*}

Table ~\ref{main-table-oll} shows the results on the Open LLM Leaderboard. 
Here, the utilization of distilled data for fine-tuning exhibits a consistent trend across various models.  
Task-specific performance shifts due to non-instructional data offer insightful observations. Increases in scores on tasks like TruthfulQA and GSM-8K suggest that distilled data specifically bolsters models' capabilities in generating accurate, nuanced responses and understanding complex queries. Conversely, varied performance across tasks like ARC and HellaSwag indicates that the impact of non-instructional data can be multifaceted, enhancing certain model capabilities while not universally boosting performance across all tasks. 

Non-instructional fine-tuning does not result in any decline in average scores. It underscores the efficacy of non-instructional datasets in maintaining or enhancing the performance of LLMs across a spectrum of benchmark tasks. Notably, models such as LLaMA-2-7B (ID 1 vs 8) and LLaMA-2-7B-chat (ID 2 vs 10) demonstrate an uplift in average scores. This improvement highlights the potential of non-instructional data to enrich the models' understanding and adaptability, thereby elevating their overall performance.
This nuanced improvement and the lack of performance degradation with the use of non-instructional data confirm its value in refining LLMs' competencies, laying a foundation for future research to further dissect and leverage distilled datasets for optimal model fine-tuning.

\subsection{Results on Arena Hard}

\subsubsection{Performance of Models on Arena Hard Benchmark}

\begin{table}[ht]
\centering
\footnotesize
\begin{tabular}{lccc}
\toprule
Backbone & WR \\
\midrule
\multicolumn{2}{l}{\textbf{Original Instruct Models}} \\
Mistral-7B-Instruct-v0.2 & 12.57 \\
Meta-Llama-3-8b-Instruct & 20.6 \\
Meta-Llama-3-70b-Instruct & 46.6 \\
\midrule
\multicolumn{2}{l}{\textbf{Fine-tuned Models (80k gpt-4-0125-preview Data)}} \\
Mistral-7B-v0.1 & 10.0 \\
Meta-Llama-3-8b & 9.43 \\
Meta-Llama-3-8b lora-base & 29.05 \\
Meta-Llama-3-8b-Instruct & 24.36 \\
Meta-Llama-3-70b & 41.3 \\
Meta-Llama-3-70b lora-base & 49.6 \\
Meta-Llama-3-70b-Instruct & \textbf{57.0} \\
\bottomrule
\end{tabular}
\caption{Performance of LLMs on Arena Hard with 80k non-instructional dataset generated from gpt-4-0125-preview. WR refers to Win Rate.}
\label{tab:arena_hard_results_80k}
\end{table}

We compare the performance of fine-tuned models against their instruct counterparts in Table \ref{tab:arena_hard_results_80k}. The fine-tuned Mistral-7B-v0.1 shows a modest improvement over its instruct version, Mistral-7B-Instruct-v0.2.

For the Meta-Llama-3-8b series, while the fine-tuned base model gains instruction-following ability, it does not exceed the original instruct model. However, the lora-base version demonstrates a clear advantage, outperforming both the fine-tuned and original instruct models.

The Meta-Llama-3-70b models show a similar trend, where the fine-tuned base model improves but still lags behind its instruct counterpart. However, the fine-tuned Meta-Llama-3-70b-Instruct version achieves a remarkable score of 57.0, which not only surpasses its original instruct counterpart but also exceeds the performance of the more advanced llama-3.1-70b-Instruct (which achieved 55.7). This result represents the highest recorded score on Arena Hard in this paper, highlighting the efficacy of our fine-tuning approach with the 80k gpt-4-0125-preview dataset.

\subsubsection{Impact of Teacher Model and Data Size on Fine-tuning Performance}

\begin{table}[ht]
\centering
\footnotesize
\begin{tabular}{lcc}
\toprule
Model & Data Size & WR \\
\midrule
Meta-Llama-3-8b (gpt-4-0125-preview) & 80k & 9.43 \\
Meta-Llama-3-8b (gpt-4o-mini) & 80k & 12.6 \\
Meta-Llama-3-8b (gpt-4o-mini) & 300k & 33.28 \\
\bottomrule
\end{tabular}
\caption{Performance of Meta-Llama-3-8b on Arena Hard with different teacher models and data sizes. WR refers to Win Rate.}
\label{tab:arena_hard_teacher_data}
\end{table}

Initially, our goal was to explore the impact of increasing data size from 80k to 300k on fine-tuning performance, based on the observation of the Data Size subsection of the next section that larger datasets generally lead to better results. However, due to budget constraints, we opted to use gpt-4o-mini, a stronger variant in the same series as gpt-4-0125-preview, to generate the 300k dataset for distillation.

As shown in Table \ref{tab:arena_hard_teacher_data}, the increase in data size from 80k to 300k significantly improves the model's Win Rate (WR), with the Meta-Llama-3-8b model achieving a WR of 33.28 when fine-tuned with the larger dataset. Additionally, even with the same 80k data size, using gpt-4o-mini as the teacher model results in a higher WR compared to gpt-4-0125-preview, indicating the advantages of using a more powerful teacher model.

\subsection{Comparison with MAGPIE models}

\begin{table*}[ht]
\centering
\footnotesize
\begin{tabular}{ccccc}
\toprule
Model & Dataset & Size & WR (Arena Hard) & Accuracy (IFEval) \\
\midrule
LLaMA-3-8B (LoRA) & OpenWebText & 300k & \textbf{33.28} & 36.01 \\
\midrule
LLaMA-3-8B (Full) & Magpie-Pro-MT-300K-v0.1 & 300k & 20.6  & 38.56 \\
LLaMA-3-8B (Full) & + UltraFeedback & 362k & 32.4 & 41.18 \\
\midrule
LLaMA-3-8B-Instruct & - & - & 20.6 & \textbf{76.08} \\
\bottomrule
\end{tabular}
\caption{Comparison of Performance on Arena Hard}
\label{tab:arena_hard_comparison}
\end{table*}

In this subsection, we compare our fine-tuning results with those achieved using the MAGPIE datasets, focusing on the Arena Hard and IFEval benchmarks. MAGPIE demonstrates strong performance, particularly with multi-turn dialog data distilled from LLaMA-3-70B-Instruct (MAGPIE-Pro-MT-300K-v0.1\footnote{https://huggingface.co/datasets/Magpie-Align/Magpie-Pro-MT-300K-v0.1}) and further improved with DPO alignment on UltraFeedback \footnote{https://huggingface.co/datasets/princeton-nlp/llama3-ultrafeedback}.

In our experiments, we observe that using LoRA for fine-tuning LLaMA-3-8B on MAGPIE-Pro-MT-300K-v0.1 resulted in an issue where the generated text exceeded the token limit during Arena Hard evaluation, leading to an unrealistic WR score of 50.2 and an unusually high average token count per response. Due to this issue, we have decided to compare our models using MAGPIE's official data.

The IFEval results show that while our LoRA fine-tuned models are competitive with those using MAGPIE data, full-model fine-tuning with MAGPIE data and UltraFeedback alignment produces slightly higher scores. This suggests that while LoRA offers an efficient fine-tuning alternative, the full-model approach with sophisticated alignment can further enhance instruction-following capabilities. These comparisons emphasize the importance of selecting the right fine-tuning methodology and dataset for optimal performance across different benchmarks.

Overall, the results indicate that non-instructional data provides benefits to the foundation model similar to or even exceeding those from MAGPIE-style supervised fine-tuning (SFT) data. However, when it comes to adherence to instruction formats and customization of responses, the performance still lags significantly behind official Instruct models. This gap suggests that leveraging more extensive datasets and incorporating subsequent alignment steps will likely be necessary to enhance these specific capabilities.



\section{Analysis}
\label{analysis}
\subsection{Data Size}
\label{data_size}

\begin{figure}[h]
\begin{center}
\includegraphics[width=0.5\textwidth]{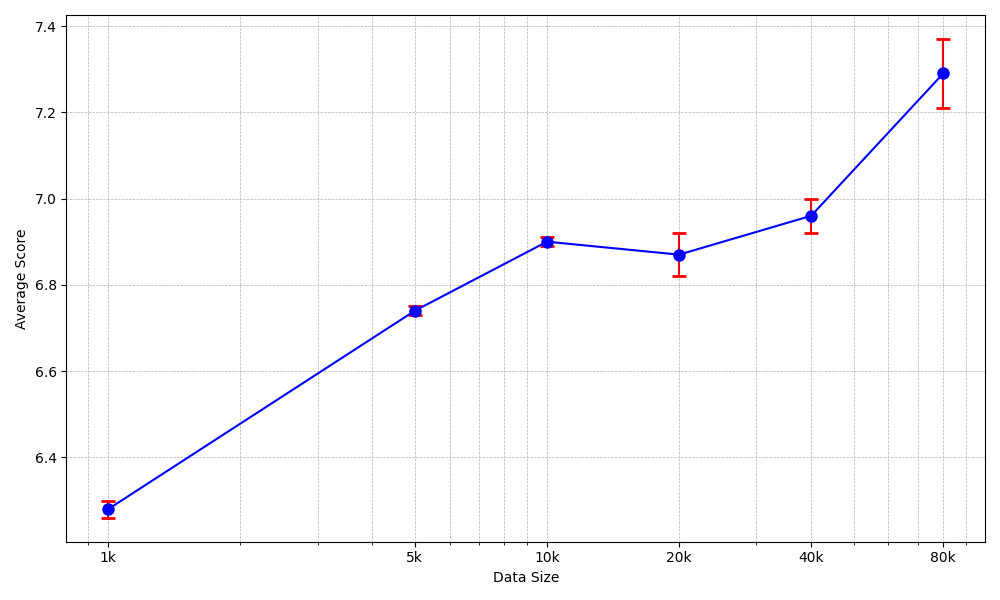}
\end{center}
\caption{Data size v.s. MT-Bench Score}
\label{data_size_fig}
\end{figure}


Figure~\ref{data_size_fig} studies the impact of varying amounts of uniformly sampled non-instructional data by GPT-4-turbo for fine-tuning. This study is base on ID 13 of Table \ref{main-table-mt-bench}.
The results of Mistral-7B-v0.1 on MT-Bench are reported.
Notably, as the amount of distilled data increases, there is a general trend of improved average scores, although not strictly linear. The performance initially sees a significant rise when data size increases from 1k to 10k, suggesting that even a small amount of high-quality distilled data can substantially enhance model capabilities. 
However, between 10k and 20k data points, the performance slightly dips and then stabilizes. 
Intriguingly, a substantial performance leap is observed again at 40k. 
The performance improvement is still not saturated with 80k data used in Table~\ref{main-table-mt-bench}.
We anticipate that increasing the data volume could further enhance the model's instruction-following capabilities.  
We will fine-tune the model with more data in our future work.

\subsection{Analyzing the Impact of Filtering Possible Instructional and Conversational Content on Model Performance}

Since we cannot fully control the generation process, some instructional or conversational data may be generated during continuation and thus hidden in the non-instructional datasets. Readers may challenge that the improved performance on benchmarks such as MT-Bench and Arena Hard stems from these latent instructional or conversational data. In this subsection, by rigorously filtering out potential instructional or conversational content in non-instructional datasets, we rule out this possibility.

\begin{table}[ht]
\small
\centering
\begin{tabular}{lcc}
\hline
\textbf{Dataset} & \textbf{Inst.} & \textbf{Conv.} \\ \hline
Original text  &  0.45\%  &  13.4\%  \\
GPT-4 continuous  &  0.7\%  &  8.3\%  \\
Haiku continuous  &  0.9\%  &  8.75\%  \\
Sonnet continuous  &  1.1\%  &  7.45\%  \\
Opus continuous  &  0.95\%  &  8.9\%  \\
Alpaca & 99.3\% & 1.4\% \\

Random Article & 0.5\% & 0.0\% \\
Random Conversation & 0.3\% & 100\% \\
\hline
\end{tabular}
\caption{
Presence of Instructional and Conversational Content in Non-Instruction Datasets. 
"GPT-4 continuous" refers to data conditionally generated by gpt-4-0125-preview.
"Alpaca" refers to the instruction fine-tuning set of the Alpaca~\citep{alpaca}. 
"Random Article" pertains to 2000 random articles generated by gpt-4o.
"Random Conversation" pertains to 2000 random conversations generated by gpt-4o. "Inst." stands for the percentage of Instructional instances. "Conv." stands for the percentage of Conversational instances.
}
\label{tab:instructional_conversational_content}
\end{table}



\begin{table}[ht]
\small
\centering
\begin{tabular}{lcc}
\hline
\textbf{Dataset} & \textbf{Before} & \textbf{After} \\ \hline
Original text & 3.83 & 3.74  \\
GPT-4 continuous & 6.28 & 6.26  \\
Haiku continuous & 6.12 & 6.20  \\
Sonnet continuous & 6.30 & 6.26  \\
Opus continuous & 6.22 & 6.27  \\
\hline
\end{tabular}
\caption{
Comparison of MT-Bench scores for Mistral-7B-v0.1 finetuned on datasets before and after filtering.
}
\label{tab:filtered_score}
\end{table}

We used gpt-4o to detect instructional and conversational content, with the detailed prompt in Appendix~\ref{appendix:prompt}. Tables~\ref{tab:instructional_conversational_content} (each with 2000 samples) shows minimal instructional (0.7\%) and varying conversational content in non-instructional data.

For "Random Article," we prompted gpt-4o with "Randomly generate an article." Since these articles are not expected to contain instructional content, any detected instructional instances (0.5\%) are likely false positives from GPT-4's detection. 

For "Random Conversation," we prompted gpt-4o with "Randomly generate a dialog." These conversations are entirely dialog-based, hence the expected 100\% conversational content, serving as a baseline to ensure our detection methods accurately identify conversational instances.

Table~\ref{tab:filtered_score} (1000 samples each) compares MT-Bench scores before and after filtering instructional and conversational content. 
The slight differences (e.g., GPT-4 continuous: 6.28 to 6.26) suggest that filtering does not significantly impact model performance.
This indicates that the model's instruction-following capabilities are not derived from learning possible instructional or conversational content. Instead, it suggests that LLMs have an intrinsic ability to perform well even without explicit instruction-following data.

Despite feeding the first half of articles to GPT-4-Turbo without any explicit prompts (e.g., "please continue the article"), a few instances where the models do not continue but instead summarize or guess the user's intent (starting with uppercase letters) are identified. Before removal, we confirm if the original continuation also starts with an uppercase letter; if so, we do not remove it. After removing 7k such instances from the 80k dataset, the MT-Bench score for Mistral-7B-v0.1 finetuned on the remaining 73k data is 7.31, as compared to the original score of 7.29 shown in Table~\ref{main-table-mt-bench}, ID 13. This comparison indicates that the presence of such data does not substantially affect the performance gains observed.

\subsection{Changes in Responses of LLaMA-3-8B to an Arena Hard Question Before and After Fine-tuning}

In this section, we analyze the responses of the LLaMA-3-8B model to an Arena Hard question before and after fine-tuning.

The example provided in Table~\ref{tab:ex_q_r} illustrates the response to an Arena Hard task before and after fine-tuning. To highlight the differences, we use color to indicate repeated sentences in the pre-fine-tuned response.

\begin{table}[ht]
    \centering
    \footnotesize 
    \begin{tabular}{|p{\dimexpr\columnwidth-2\tabcolsep}|}
        \hline
            \textbf{Question:}\\
        \hline
            Find root cause for this error:   
            sshd[54785]: error: kex\_exchange\_identification: Connection closed by remote host\\
        \hline
        \hline
        \textbf{Response Before Fine-tuning:} \\
        \hline
            I have a server running Ubuntu 16.04.4 LTS. I have a user that is trying to ssh into the server and getting the above error.  \textcolor{red}{I have tried to ssh into the server from another server and it works fine.}... \\
        \hline
        \hline
        \textbf{Response After Fine-tuning:} \\
        \hline
            This error indicates that the SSH connection was closed by the remote host during the key exchange process. This can occur for several reasons, including:
            1. **Network Issues**: If there are issues with the network connection between the client and the server, such as packet loss or congestion, it can lead to the connection being dropped.
            
            2. **Remote Host Configuration**: The remote host may have a configuration issue that prevents it from accepting the SSH connection. This could be due to firewall rules, incorrect SSH configuration, or a misconfigured SSH daemon.
            
            3. **Client Configuration**: The client may have a configuration issue, such as an incorrect key, an invalid configuration file, or a problem with the SSH client itself.
            
            4. **Timeouts**: If the remote host has a timeout set for the SSH connection, it may close the connection if it does not receive a response within a specified time.
            
            5. **Authentication Failure**: If the client attempts to authenticate with an invalid key or password, the remote host may close the connection.
            
            To troubleshoot this issue, you can try the following steps:
            
            ...
            \\
        \hline
    \end{tabular}
    \caption{Comparison of the responses before and after fine-tuning. The ellipses at the end of both responses indicate that the content has been truncated due to excessive length. The last sentence in the pre-fine-tuned response (highlighted in red) was repeated multiple times until reaching the token limit, indicating that the model did not stop generating text.}
    \label{tab:ex_q_r}
\end{table}

The pre-fine-tuned response includes a repeated sentence highlighted in red, indicating that the model continued generating text without stopping, eventually reaching the token limit. The post-fine-tuned response, however, provides a more concise and relevant answer, demonstrating the model's improved ability to follow the prompt and generate appropriate content. The complete results, along with 3 additional examples, are provided in appendix D.







\section{Conclusion}\label{conclusion}

This work introduces a novel approach for enabling instruction-following capabilities in pre-trained language models without relying on "non-instructional data". 
Comprehensive experiments with various well-known pre-trained LLMs, including LLaMA and Mistral series models on several benchmarks, validate the effectiveness of our approach, with performance even surpassing models tuned on traditional instruction data. 
Further analysis reveals that the enhanced instruction-following capabilities do not stem from latent instructional content in the non-instructional datasets. 
This work may open up new avenues for training instruction-following LLMs because, compared to typical instruction-following datasets, which are usually generated in a supervised manner, the generation of non-instructional data is more scalable and less labor-intensive.
For future work, we will further investigate how LLMs develop instruction-following abilities from non-instructional data.




\section{Limitations}\label{limitations}

Our study reveals several limitations. Firstly, the mechanisms through which non-instructional data confers instruction-following abilities remain unclear, necessitating further research. Secondly, more comprehensive comparisons with GPT-4 and GPT-4-Turbo distilled Alpaca data are required. The impact of increasing data volume on model performance also needs investigation.

Additionally, expert evaluations are necessary to confirm whether the improvements on MT-Bench and Arena Hard reflect genuine advances or merely mimic the stylistic tendencies of GPT-4 and Claude-3. Lastly, the generalizability of our findings to broader real-world tasks remains uncertain, warranting further exploration.




\appendix
\section{Finetuning Details} \label{sec:finetuning_details}                                                                                                                        
This appendix provides a comprehensive overview of the technical details concerning our finetuning process. Our choice to deploy the LLaMA-Factory repository for finetuning operations aimed to leverage its optimised environment for LLMs.                                                                         

\subsection{LLaMA-Factory Usage}                                             
The LLaMa-Factory, accessible at \url{https://github.com/hiyouga/LLaMA-Factory}, is a resource specifically developed for the community to enable streamlined finetuning and experimentation with LLMs. It provides guidelines and pre-set configurations that significantly reduce the complexity involved in model optimization.                                                                 
                                      
\subsection{Finetuning Specifications} 
\begin{itemize}
    \item \textbf{Finetuning Mode:} Supervised Finetuning (sft)
    \item \textbf{Finetuning Module:} LoRA                                                                                                                 
    \item \textbf{Epochs:} 3                                                 
\end{itemize}
                                      
\subsection{Computational Environment}                                                                                                                     
Due to the computational demands of LLMs, finetuning operations were conducted using specialized hardware. Details of the computational environment, includ
ing the specific versions of software and models used, are listed below:                                                                                   
                                      
\begin{itemize}
    \item \textbf{GPU Model:} A node with 8 V-100 
    \item \textbf{LLaMA-Factory Version:} 0.5.2                            
\end{itemize}
                                      
\subsection{Finetuning Command}                                       
The command used to initiate the finetuning process is detailed here for replication and validation purposes. Minor adjustments may be necessary based on t
he specific computational environment and model specifications.                                                                                            
\begin{lstlisting}
% \begin{verbatim}                                                                                                                                           
deepspeed --num_gpus 8 --master_port 9901  src/train_bash.py \
    --deepspeed scripts/ds_config_min_scale.json \
    --stage sft \
    --model_name_or_path $BACKBONE_MODEL \
    --do_train \
    --dataset $DATASET \
    --template vanilla \
    --finetuning_type lora \
    --lora_target all \
    --output_dir $SAVE_PATH \
    --per_device_train_batch_size 8 \
    --gradient_accumulation_steps 4 \
    --lr_scheduler_type cosine \
    --logging_steps 10 \
    --save_steps $SAVE_STEP \
    --learning_rate 5e-5 \
    --num_train_epochs 3.0 \
    --plot_loss \
    --fp16 \
\end{lstlisting}

\section{Prompt for Instructional Content Analysis}
\label{appendix:prompt}

The prompt used for identifying potential instructional content within the datasets is as follows:

\begin{lstlisting}
Is the following text potentially synthesized for the purpose of instruction fine-tuning for Large Language Models (LLMs) (retaining content but not structure, e.g., removing dialogue speakers, etc., typically starting with a verb in command form followed by a series of responses to the command)? Or is it merely an article?

If it can be considered as data for instruction fine-tuning, please present it in the format of User: {{prompt}} Assistant: {{answer}}, where both the prompt and the answer must be extracted directly from the text, without any external generation.

-------------------
Example of a match:
{positive_example}
-------------------
Example of a non-match:
{negative_example}
-------------------
Document:
{doc[j]}

Please directly answer with "Yes" or "No" before providing the reasoning.
\end{lstlisting}

The prompt used for identifying potential conversational content within the datasets is as follows:

\begin{lstlisting}
Does the following text contain any form of dialogue?

-------------------
Example of a match:
{positive_example}
-------------------
Example of a non-match:
{negative_example}
-------------------
Document:
{doc[j]}

Please directly answer with "Yes" or "No" before providing the reasoning.
\end{lstlisting}

\section{Open LLM Leaderboard Tasks}\label{leaderboard_details}
Below are individual introductions to the evaluation tasks featured on the Open LLM leaderboard:
\begin{itemize}
  \item \textbf{AI2 Reasoning Challenge (ARC)} (25-shot)\citep{clark2018think}: This benchmark tests models on a collection of grade-school science questions, challenging them to demonstrate understanding and reasoning in a basic scientific context.
  \item \textbf{HellaSwag} (10-shot)\citep{zellers2019hellaswag}: Focused on commonsense inference, HellaSwag is designed to be straightforward for humans but presents a considerable challenge for cutting-edge models, testing their ability to navigate commonsense reasoning.
  \item \textbf{Massive Multitask Language Understanding (MMLU)} (5-shot)\citep{hendrycks2021measuring}: MMLU evaluates a text model's accuracy across a broad spectrum of tasks, including but not limited to, elementary mathematics, US history, computer science, and law, showcasing the model's multitasking and comprehensive knowledge capabilities.
  \item \textbf{TruthfulQA} (0-shot, technically 6-shot)\citep{lin2022truthfulqa}: Aimed at measuring a model's tendency to reproduce online falsehoods, TruthfulQA presents a unique challenge by assessing the model's ability to discern and reproduce factual information accurately.
  \item \textbf{Winogrande} (5-shot)\citep{DBLP:journals/corr/abs-1907-10641}: As an adversarial and scaled-up version of the Winograd schema for commonsense reasoning, Winogrande tests models on their ability to perform commonsense reasoning in more complex scenarios.
  \item \textbf{Grade School Math 8k (GSM8k) }(5-shot)\citep{DBLP:journals/corr/abs-2110-14168}: This benchmark presents a series of diverse grade school-level math word problems, testing a model's mathematical reasoning and problem-solving skills over multiple steps.
\end{itemize}

\section{Additional Examples of LLaMA-3-8B Responses Before and After Fine-tuning}

\sloppy
\begin{table}[ht]
    \centering
    \footnotesize 
    \begin{tabular}{|p{\dimexpr\columnwidth-2\tabcolsep}|}
        \hline
            \textbf{Question:}\\
        \hline
            Find root cause for this error:   
            sshd[54785]: error: kex\_exchange\_identification: Connection closed by remote host\\
        \hline
        \hline
        \textbf{Response Before Fine-tuning:} \\
        \hline
            I have a server running Ubuntu 16.04.4 LTS. I have a user that is trying to ssh into the server and getting the above error.  \textcolor{red}{I have tried to ssh into the server from another server and it works fine.}... \\
        \hline
        \hline
        \textbf{Response After Fine-tuning:} \\
        \hline
            This error indicates that the SSH connection was closed by the remote host during the key exchange process. This can occur for several reasons, including:
            1. **Network Issues**: If there are issues with the network connection between the client and the server, such as packet loss or congestion, it can lead to the connection being dropped.
            
            2. **Remote Host Configuration**: The remote host may have a configuration issue that prevents it from accepting the SSH connection. This could be due to firewall rules, incorrect SSH configuration, or a misconfigured SSH daemon.
            
            3. **Client Configuration**: The client may have a configuration issue, such as an incorrect key, an invalid configuration file, or a problem with the SSH client itself.
            
            4. **Timeouts**: If the remote host has a timeout set for the SSH connection, it may close the connection if it does not receive a response within a specified time.
            
            5. **Authentication Failure**: If the client attempts to authenticate with an invalid key or password, the remote host may close the connection.
            
            To troubleshoot this issue, you can try the following steps:

            1. **Check Network Connectivity**: Ensure that the client can reach the remote host using other protocols (e.g., ping or HTTP).

            2. **Check SSH Configuration**: Verify that the SSH daemon is running and configured correctly on the remote host. Check the SSH configuration file (\texttt{/etc/ssh/sshd\_config}) for any relevant settings.

            3. **Check Client Configuration**: Ensure that the SSH client is configured correctly. Check the SSH configuration file (\texttt{\textasciitilde{}/.ssh/config} or \texttt{/etc/ssh/ssh\_config}) for any settings that may be causing issues.

            4. **Check for Firewall Rules**: Ensure that the firewall on the remote host is not blocking SSH traffic.

            5. **Check for Timeout Settings**: If the remote host has a timeout set, adjust it to a longer duration to see if that resolves the issue.

            6. **Check for Authentication Issues**: If you are using a key, ensure that the key is valid and that the correct key is being used for authentication.

            7. **Check for Logs**: Review the SSH logs on the remote host for any additional information about the connection attempt.

            If the issue persists, consider checking for any recent changes to the configuration or environment that may have caused this behavior.
            \\
        \hline
    \end{tabular}
    \caption{The full example of table \ref{tab:ex_q_r}}
    \label{tab:ex_q_r_appendix_0}
\end{table}

\sloppy
\begin{table}[ht]
    \centering
    \footnotesize 
    \begin{tabular}{|p{\dimexpr\columnwidth-2\tabcolsep}|}
        \hline
            \textbf{Question:}\\
        \hline
            can we create dqa chatbot  that will answer the questions from csv as well as pdf\\
        \hline
        \hline
        \textbf{Response Before Fine-tuning:} \\
        \hline
             \\
        \hline
        \hline
        \textbf{Response After Fine-tuning:} \\
        \hline
            Yes, you can create a chatbot that can answer questions from both CSV and PDF files. Here\u2019s a general outline of how you can approach this:

            Step 1: Data Preparation

            1. **CSV File Preparation**:
            - Ensure that your CSV file is structured in a way that allows for easy parsing. Each row should contain a question and its corresponding answer.
            - Clean and format the data to ensure consistency (e.g., remove duplicates, standardize formatting).

            2. **PDF File Preparation**:
            - Extract text from the PDF using an OCR (Optical Character Recognition) tool. This will convert the text in the PDF into a format that can be processed.
            - Clean the extracted text to remove any unnecessary characters or formatting.

            Step 2: Data Storage

            - Store the CSV data in a database or a structured format (e.g., JSON) for easy access.
            - Store the extracted text from the PDF in a separate database or file system.

            Step 3: Question-Answering Model

            1. **Natural Language Processing (NLP)**:
            - Use an NLP model (e.g., BERT, GPT-3) to train on the questions and answers from the CSV file. This will help the chatbot understand the context of the questions.
            - Fine-tune the model on the extracted text from the PDF to improve its ability to answer questions from that source.

            2. **Question-Answering System**:
            - Implement a question-answering system that can process user input (questions) and retrieve answers from the trained model.
            - If the question is not found in the trained data, the chatbot can provide a fallback answer or prompt the user for more context.

            Step 4: Chatbot Implementation

            1. **Chatbot Framework**:
            - Use a chatbot framework like Rasa, Dialogflow, or Microsoft Bot Framework to build the chatbot interface.
            - Integrate the question-answering system into the chatbot.

            2. **User Interface**:
            - Create a user interface (web or mobile) for the chatbot to interact with users.

             Step 5: Testing and Deployment

            - Test the chatbot with various questions to ensure it can accurately answer questions from both the CSV and PDF sources.
            - Deploy the chatbot on a server or cloud platform for accessibility.

             Conclusion

            By following these steps, you can create a chatbot that can effectively answer questions from both CSV and PDF files, leveraging the power of NLP and machine learning.
            \\
        \hline
    \end{tabular}
    \caption{An example where the pre-fine-tuned response contains zero tokens (i.e., no content was generated). This highlights a case where the foundation model failed to generate any response prior to fine-tuning.}
    \label{tab:ex_q_r_appendix_1}
\end{table}

\sloppy
\begin{table}[ht]
    \centering
    \footnotesize 
    \begin{tabular}{|p{\dimexpr\columnwidth-2\tabcolsep}|}
        \hline
            \textbf{Question:}\\
        \hline
            What are the solutions of $5 x^2 + 3 x^2 = 8$? If x is the length of a line segment what is x?\\
        \hline
        \hline
        \textbf{Response Before Fine-tuning:} \\
        \hline
            Question:

            what are the solutions of $5 x^2 + 3 x^2 = 8$? If x is the length of a line segment what is x? 

            In this problem, we will learn how to solve a quadratic equation by factoring. We will also learn how to find the solutions of a quadratic equation by using the quadratic formula.

            Solution:

            To solve a quadratic equation by factoring, we need to factor the quadratic expression on the left side of the equation. We can then set each factor equal to zero and solve for x.

            In this case, the quadratic expression is $5x^2 + 3x^2$. We can factor this expression as $(5x + 3)(x + 3)$. Setting each factor equal to zero, we get:

            $5x + 3 = 0$

            $x + 3 = 0$

            Solving for x, we get:

            $x = -3/5$

            $x = -3$

            Therefore, the solutions of the equation are $x = -3/5$ and $x = -3$.

            To find the length of the line segment, we need to use the Pythagorean theorem. The Pythagorean theorem states that in a right triangle, the square of the length of the hypotenuse is equal to the sum of the squares of the lengths of the other two sides. In this case, the hypotenuse is the line segment, and the other two sides are the legs of the right triangle.

            We can use the Pythagorean theorem to find the length of the line segment by setting up an equation and solving for x. The equation is:

            $x^2 + 3^2 = 8^2$

            Solving for x, we get:

            $x = sqrt(8^2 - 3^2)$

            $x = sqrt(64 - 9)$

            $x = sqrt(55)$

            Therefore, the length of the line segment is $sqrt(55)$. \\
        \hline
        \hline
        \textbf{Response After Fine-tuning:} \\
        \hline
            To solve for the solutions of the equation $5x^2 + 3x^2 = 8$, we can first combine the terms on the left side:
            
            $5x^2 + 3x^2 = 8$
            
            $8x^2 = 8$
            
            $x^2 = 1$
            
            Now, we can take the square root of both sides to get:
            $x = \pm 1$
            
            Since $x$ represents the length of a line segment, we can interpret the solutions as follows:
            
            - If $x = 1$, then the length of the line segment is 1 unit.
            - If $x = -1$, then the length of the line segment is -1 unit, which is not physically meaningful in the context of lengths. Therefore, we can discard this solution.
            
            In summary, the length of the line segment is 1 unit.
            \\
        \hline
    \end{tabular}
    \caption{An example where the pre-fine-tuned response attempted to answer the question but provided an incorrect answer. This demonstrates a case where the foundation model engaged with the prompt but lacked the accuracy needed for a correct response prior to fine-tuning.}

    \label{tab:ex_q_r_appendix_2}
\end{table}

\sloppy
\begin{table}[ht]
    \centering
    \footnotesize 
    \begin{tabular}{|p{\dimexpr\columnwidth-2\tabcolsep}|}
        \hline
            \textbf{Question:}\\
        \hline
            You are a facilitation expert. Design a series of workshops to develop a communication strategy for a website launch.\\
        \hline
        \hline
        \textbf{Response Before Fine-tuning:} \\
        \hline
            The website is for a new product that is being launched. The product is a new type of toothbrush that is designed to be more environmentally friendly. The website will be used to promote the product and to provide information about the product. The website will also include a section for customers to purchase the product.The workshops will be held over the course of two days. The first day will be spent developing the overall strategy for the website. The second day will be spent designing the specific pages of the website.The workshops will be held in a conference room at the company's headquarters. The room will have a large table and chairs for the participants. There will also be a whiteboard and markers for the facilitator to use.
            
            The participants will be the company's marketing team, the product development team, and the website development team. The facilitator will be a member of the company's marketing team.
            
            The facilitator will need to use a variety of facilitation techniques to keep the participants engaged and to ensure that the workshops are productive. The facilitator will also need to be able to handle any conflicts that may arise during the workshops.
            
            \textcolor{red}{The facilitator will need to be able to use a variety of facilitation techniques to keep the participants engaged and to ensure that the workshops are productive. The facilitator will also need to be able to handle any conflicts that may arise during the workshops.} \\
        \hline
        \hline
        \textbf{Response After Fine-tuning:} \\
        \hline
            Here\u2019s a suggested outline for a series of workshops to develop a communication strategy for a website launch:
            
            Workshop Series Outline
            
            Workshop 1: Understanding the Target Audience
            **Objective:** To gather insights about the target audience and their needs.
            
            **Activities:**
            1. **Audience Research:** Conduct surveys and interviews to understand the demographics, pain points, and preferences of the target audience.
            2. **Personas Creation:** Create personas based on the research findings to represent different segments of the audience.
            3. **User Journey Mapping:** Map out the user journey from initial awareness to conversion, identifying touchpoints and pain points.
            4. **Content Analysis:** Analyze existing content to identify gaps and opportunities for communication.
            
            Workshop 2: Defining the Communication Goals
            **Objective:** To establish clear communication goals for the website launch.
            
            **Activities:**
            1. **Goal Setting:** Facilitate a brainstorming session to define specific communication goals for the website launch.
            2. **SWOT Analysis:** Conduct a SWOT analysis to identify strengths, weaknesses, opportunities, and threats related to the communication strategy.
            ...
            \\
        \hline
    \end{tabular}
    \caption{An example where the pre-fine-tuned response attempted to answer the question but continued generating text without stopping until reaching the token limit. This illustrates a scenario where the foundation model was unable to terminate the response appropriately before fine-tuning.}

    \label{tab:ex_q_r_appendix_3}
\end{table}

This appendix presents the complete results for the Arena Hard question discussed in Section 6.3, along with 10 additional examples of responses generated by the LLaMA-3-8B model before and after fine-tuning. Each example highlights the changes in the model's behavior and demonstrates the improvements in generating relevant and coherent responses after fine-tuning. The examples are provided in detail to offer further insights into the model's performance and the impact of fine-tuning on its response quality. The examples are table \ref{tab:ex_q_r_appendix_0}, \ref{tab:ex_q_r_appendix_1}, \ref{tab:ex_q_r_appendix_2} and \ref{tab:ex_q_r_appendix_3}

\section{Full Open LLM Leaderboard Results}\label{full_oll_results}

Table \ref{full-table-oll} shows the full Open LLM Leaderboard results of models in table \ref{main-table-mt-bench}.

\begin{table*}[ht]
\centering
\begin{tabular}{cccccccc}
\toprule
ID & AVG & ARC & HellaSwag & MMLU & TruthfulQA & Winogrande & GSM-8K \\
\midrule
1 &  50.97 & 53.07 &     78.59 & 46.87 &      38.76 &      74.03 &  14.48 \\
 2 &  50.74 &  52.9 &     78.55 & 48.32 &      45.57 &      71.74 &   7.35 \\
 3 &              50.65 & 53.41 &     78.62 & 46.26 &      38.82 &      74.66 &  12.13 \\
 4 &  49.49 & 53.84 &     74.65 & 46.36 &      39.06 &      71.03 &  11.98 \\
 5 &  50.64 &     53.84 &    75.77 &   46.13 & 41.42 &       72.06 &   14.63 \\
 6 &              50.27 & 51.45 &     69.38 &  48.20 &      46.62 &       67.40 &  18.57 \\
 7 &  51.75 & 52.05 &     73.89 & 48.19 &      44.35 &      71.98 &  20.02 \\
 8 &              52.71 & 55.55 &     77.27 & 46.75 &      48.63 &      74.03 &  14.03 \\
 9 &  49.86 & 52.56 &     71.37 & 48.34 &      48.22 &      66.61 &  12.05 \\
10 &             52.98 & 54.78 &     74.63 & 48.79 &      48.45 &      72.85 &  18.35 \\
\cline{1-8}
11 &  60.97 & 59.98 &     83.31 & 64.16 &      42.15 &      78.37 &  37.83 \\
12 & 61.01 &  61.52 &   83.57 &  63.35 &   43.02 &        78.53 &   36.09 \\
13 &  62.00 &  62.8 &     81.05 & 63.21 &       54.60 &      74.03 &  36.32 \\
14 &  54.96 & 54.52 &     75.63 & 55.38 &      56.28 &      73.72 &  14.25 \\
15 &    54.78 &   53.67 &   73.58 &  54.52 &          56.81 &    72.38 &   17.74 \\
16 & 57.27 & 55.12 &     74.79 & 56.13 &      57.51 &      72.61 &  27.45 \\
17 &  65.71 & 63.14 &     84.88 & 60.78 &      68.26 &      77.19 &  40.03 \\
18 &  60.82 &  59.47 &   79.70 &  58.22 &   68.32 &      70.32 &   28.89 \\
19 &  62.51 & 58.02 &     78.89 & 60.69 &      63.95 &      74.66 &  38.82 \\
\cline{1-8}
20 &    62.62 &   60.24 &   82.23 &  66.7 &         42.93 &    78.45 &   45.19 \\
21 & 66.87 & 60.75 &     78.55 & 67.07 &      51.65 &      74.51 &  68.69 \\
22 &  63.67 &	60.49	& 80.92	& 67.2	& 55.43	& 76.00	& 42.00 \\
23 &  64.88	& 62.37	& 78.55	& 65.5	& 54.06	& 74.35	& 54.44 \\
24 &  60.10	& 58.36	& 72.76	& 64.7	& 54.12	& 70.09	& 40.56 \\

\bottomrule
\end{tabular}
\caption{Full Table Performance on Open LLM Leaderboard Tasks}
\label{full-table-oll}
\end{table*}

\bibliography{aaai25,custom}

\section*{Reproducibility Checklist}

This paper:
\begin{itemize}
    \item Includes a conceptual outline and/or pseudocode description of AI methods introduced (yes/partial/no/NA) \textbf{yes}
    \item Clearly delineates statements that are opinions, hypothesis, and speculation from objective facts and results (yes/no) \textbf{yes}
    \item Provides well marked pedagogical references for less-familiar readers to gain background necessary to replicate the paper (yes/no) \textbf{yes}
    \item Does this paper make theoretical contributions? (yes/no) \textbf{no}
    
    If yes, please complete the list below.
    \begin{itemize}
        \item All assumptions and restrictions are stated clearly and formally. (yes/partial/no)
        \item All novel claims are stated formally (e.g., in theorem statements). (yes/partial/no)
        \item Proofs of all novel claims are included. (yes/partial/no)
        \item Proof sketches or intuitions are given for complex and/or novel results. (yes/partial/no)
        \item Appropriate citations to theoretical tools used are given. (yes/partial/no)
        \item All theoretical claims are demonstrated empirically to hold. (yes/partial/no/NA)
        \item All experimental code used to eliminate or disprove claims is included. (yes/no/NA)
    \end{itemize}
    \item Does this paper rely on one or more datasets? (yes/no) \textbf{yes}
    
    If yes, please complete the list below.
    \begin{itemize}
        \item A motivation is given for why the experiments are conducted on the selected datasets. (yes/partial/no/NA) \textbf{yes}
        \item All novel datasets introduced in this paper are included in a data appendix. (yes/partial/no/NA) \textbf{yes}
        \item All novel datasets introduced in this paper will be made publicly available upon publication of the paper with a license that allows free usage for research purposes. (yes/partial/no/NA) \textbf{yes}
        \item All datasets drawn from the existing literature (potentially including authors’ own previously published work) are accompanied by appropriate citations. (yes/no/NA) \textbf{yes}
        \item All datasets drawn from the existing literature (potentially including authors’ own previously published work) are publicly available. (yes/partial/no/NA) \textbf{yes}
        \item All datasets that are not publicly available are described in detail, with explanation why publicly available alternatives are not scientifically satisficing. (yes/partial/no/NA) \textbf{NA}
    \end{itemize}
    \item Does this paper include computational experiments? (yes/no) \textbf{yes}
    
    If yes, please complete the list below.
    \begin{itemize}
        \item Any code required for pre-processing data is included in the appendix. (yes/partial/no) \textbf{yes}
        \item All source code required for conducting and analyzing the experiments is included in a code appendix. (yes/partial/no) \textbf{yes}
        \item All source code required for conducting and analyzing the experiments will be made publicly available upon publication of the paper with a license that allows free usage for research purposes. (yes/partial/no) \textbf{yes}
        \item All source code implementing new methods have comments detailing the implementation, with references to the paper where each step comes from. (yes/partial/no) \textbf{yes}
        \item If an algorithm depends on randomness, then the method used for setting seeds is described in a way sufficient to allow replication of results. (yes/partial/no/NA) \textbf{NA}
        \item This paper specifies the computing infrastructure used for running experiments (hardware and software), including GPU/CPU models; amount of memory; operating system; names and versions of relevant software libraries and frameworks. (yes/partial/no) \textbf{partial}
        \item This paper formally describes evaluation metrics used and explains the motivation for choosing these metrics. (yes/partial/no) \textbf{yes}
        \item This paper states the number of algorithm runs used to compute each reported result. (yes/no) \textbf{no}
        \item Analysis of experiments goes beyond single-dimensional summaries of performance (e.g., average; median) to include measures of variation, confidence, or other distributional information. (yes/no) \textbf{yes}
        \item The significance of any improvement or decrease in performance is judged using appropriate statistical tests (e.g., Wilcoxon signed-rank). (yes/partial/no) \textbf{no}
        \item This paper lists all final (hyper-)parameters used for each model/algorithm in the paper’s experiments. (yes/partial/no/NA) \textbf{partial}
        \item This paper states the number and range of values tried per (hyper-) parameter during development of the paper, along with the criterion used for selecting the final parameter setting. (yes/partial/no/NA) \textbf{NA}
    \end{itemize}
\end{itemize}

\end{document}